\documentclass[lettersize,journal]{IEEEtran}
\usepackage{amsmath,amsfonts}
\usepackage{algorithmic}
\usepackage{algorithm}
\usepackage{array}
\usepackage[caption=false,font=normalsize,labelfont=sf,textfont=sf]{subfig}
\usepackage{textcomp}
\usepackage{stfloats}
\usepackage{url}
\usepackage{verbatim}
\usepackage{graphicx}
\usepackage{cite}
\usepackage{hyperref}
\usepackage{color}
\begin{document}

\title{An Efficient LiDAR-Camera Fusion Network for Multi-Class 3D Dynamic Object Detection and Trajectory Prediction}

% \author{Anonymous submission}
\author{Yushen He$^{1}$, Lei Zhao$^{1}$, Tianchen Deng$^{1}$, Zipeng Fang$^{1}$, and Weidong Chen$^{1*}$, \IEEEmembership{Member, IEEE} % <-this % stops a space
\thanks{*The corresponding author, wdchen@sjtu.edu.cn.}% <-this % stops a space
\thanks{$^{1}$Institute of Medical Robotics and Department of Automation, Shanghai Jiao Tong University, Key Laboratory of System Control and Information Processing, Ministry of Education, Shanghai 200240, China.}%
}



\maketitle

\begin{abstract}
Service mobile robots are often required to avoid dynamic objects while performing their tasks, but they usually have only limited computational resources. 
To further advance the practical application of service robots in complex dynamic environments, we propose an efficient multi-modal framework for 3D object detection and trajectory prediction, which synergistically integrates LiDAR and camera inputs to achieve real-time perception of pedestrians, vehicles, and riders in 3D space.
The framework incorporates two novel models: 1) a Unified modality detector with Mamba and Transformer (UniMT) for object detection, which achieves high-accuracy object detection with fast inference speed, 
and 2) a Reference Trajectory-based Multi-Class Transformer (RTMCT) for efficient and diverse trajectory prediction of multi-class objects with flexible-length trajectories. 
Evaluations on the CODa benchmark demonstrate that our method outperforms existing ones in both detection (+3.71\% in mAP) and trajectory prediction (-0.408m in minADE$_5$ of pedestrians) metrics. 
Furthermore, on the challenging nuScenes detection benchmark, our detection model achieves competitive performance among LiDAR-camera fusion methods, with a mAP of 72.7\% and NDS of 75.3\%. 
Remarkably, the system demonstrates exceptional generalizability and practical deployment potential. When transferred and implemented on a wheelchair robot with an entry-level NVIDIA RTX 3060 GPU, it achieves real-time inference at 13.9 frames per second (FPS) with satisfactory accuracy. 
To facilitate reproducibility and practical deployment, we release the related code of the method at \href{https://github.com/TossherO/3D_Perception}{https://github.com/TossherO/3D\_Perception} and its ROS inference version at \href{https://github.com/TossherO/ros_packages}{https://github.com/TossherO/ros\_packages}.
\end{abstract}

\begin{IEEEkeywords}
Object Detection, Trajectory Prediction, Sensor Fusion, Mobile Robots
\end{IEEEkeywords}

\section{INTRODUCTION}

\IEEEPARstart{S}{afe} autonomous navigation for service mobile robots in daily scenarios fundamentally relies on robust 3D scene understanding \cite{SceneRepresentation,sfpnet} and the accurate perception of dynamic objects such as pedestrians, vehicles, and riders. 
Existing approaches typically adopt end-to-end mode or modular mode (sequential models of detection, tracking, and trajectory prediction) to perform the perception tasks. 
While end-to-end models demonstrate competitive performance, their high computational demands for both training and inference, as well as scene-specific optimization (typically tailored for autonomous driving \cite{auto1,auto2}), render them impractical for resource-limited mobile robots operating in diverse daily environments. 
In contrast, the modular sequential framework facilitates real-time deployment through independent optimization of its components, which better aligns with the operational constraints of mobile robots. 
The subsequent discussion will focus on the three components of the sequential framework: object detection, object tracking, and trajectory prediction.

\begin{figure}
        \centering
        \includegraphics[width=3.4in]{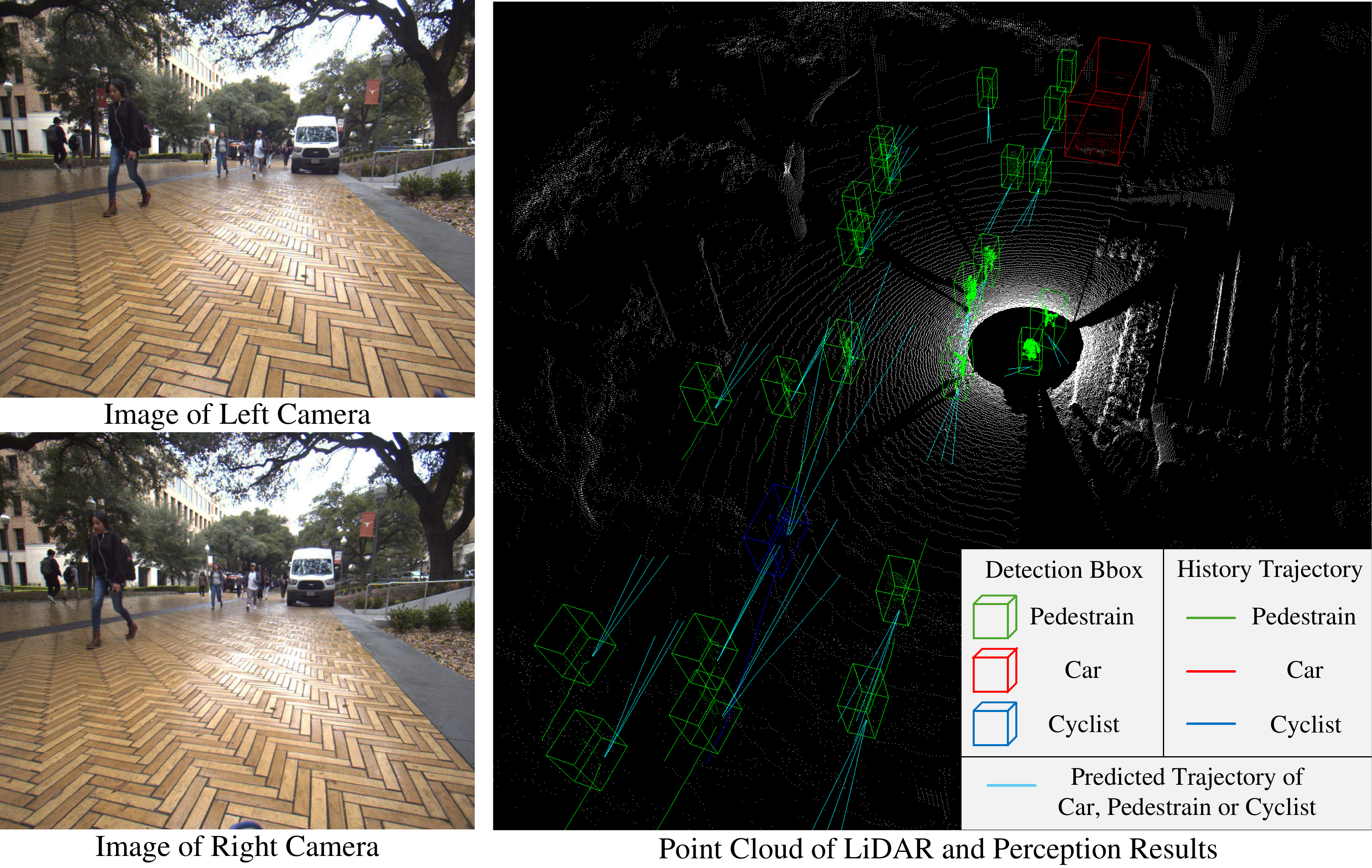}
        \caption{A qualitative example of our method's 3D detection and multi-class trajectory prediction results on the CODa dataset.}
        \label{finalresult}
\end{figure}

\textbf{The detection model}, which processes raw sensor inputs to generate 3D bounding boxes, constitutes the most critical component of the system. 
Performance when using LiDAR or cameras individually is limited and may encounter problems in many scenarios. 
LiDAR-camera fused detection models tend to be high-performing, but often exhibit complex structures. 
Currently, there are many fusion strategies between point clouds and images. 
Some methods \cite{pointpainting,virconv} convert image pixels to 3D virtual points, enabling early fusion at the point level. 
Other methods \cite{bevfusion} transform image features into 3D space by Lift-Splat-Shot (LSS) \cite{lss}, then align and merge them with point cloud features to enable middle fusion at the feature level. 
CMT \cite{cmt} employs a DETR-like \cite{detr} decoder, which uses sparse queries and global attention to enable implicit post fusion. 
However, these methods all have some drawbacks. 
In terms of fusion accuracy, point-level fusion and feature-level fusion are vulnerable to errors in image depth estimation and the LiDAR-camera transformation matrix. 
While implicit fusion of CMT exhibits greater robustness, its global attention struggles to extract precise object information. 
In terms of computational complexity, point-level fusion requires pre-processing for depth estimation, and feature-level fusion involves transformation from 2D image features to 3D space in LSS, both of which are time-consuming. 
The global attention in implicit fusion also entails substantial computational and memory overhead when processing large feature maps. 
Consequently, there arises a critical need for detection model that achieve both high accuracy and computational efficiency.

\textbf{The tracking model} utilizes temporal detection results to obtain object trajectories, building a bridge between detection and trajectory prediction tasks. 
We adopt the non-learning tracking-by-detection (TBD) method SimpleTrack \cite{simpletrack} as our tracking model due to its lightweight nature and competitive performance. 
To further accelerate inference, we reimplemented it on the GPU.

\textbf{The trajectory prediction model} predicts the possible future trajectories of current observed objects based on historical information. 
Some works \cite{vectornet,lanercnn}, primarily designed for autonomous driving scenarios, rely on map priors, incorporate traffic rules, and focus on modeling lane structures. 
They are unsuitable for mobile robots operating in daily scenarios, which typically lack well-defined maps, traffic rules, and lane markings. 
Other works \cite{sociallstm,stgat,star,desire,socialvae,socialimplicit,socialgan,sophie} focus on pedestrian trajectory prediction. 
These methods typically require fixed-length historical trajectories as input and are unable to handle multiple object classes. 
Moreover, to achieve diversity in future trajectories, complex generative models are often employed. 
However, the history trajectories provided by the tracking model have flexible lengths and multiple classes. 
It becomes imperative to develop approaches that can handle flexible-length, multi-class trajectories using efficient architectures for diverse trajectory prediction.

To address the aforementioned challenges and meet the requirements for safe autonomous navigation in daily environments, we develop an efficient perception system for service mobile robots. 
This system utilizes LiDAR and camera sensors to perform real-time 3D detection, tracking, and trajectory prediction for multiple classes of dynamic objects, including pedestrians, vehicles, and riders. 

Our contributions are mainly as follows:

\begin{itemize}
\item We propose an efficient multi-modal framework for 3D object detection and trajectory prediction. 
This framework integrates our novel detection and trajectory prediction models with the GPU-accelerated SimpleTrack for object tracking.
\item We propose the efficient LiDAR-camera fused detection model named Unified modality detector with Mamba and Transformer (UniMT). 
We introduce the Multi-model Mamba Encoder (MME) to process multi-modal features with deep yet soft fusion strategies and efficient Mamba blocks. 
Additionally, we introduce the 3D Multi-model Deformable Attention (MDA) module used in the DETR-like decoder to extract sparse and precise multi-modal features based on 3D queries.
\item We propose the trajectory prediction model named Reference Trajectory based Multi-Class Transformer (RTMCT). 
It can predict future trajectories for multiple classes of objects with flexible trajectory lengths by reference trajectories and Transformer, while achieving high computational efficiency and diversity of predicted trajectories.
\item Extensive evaluations on the dataset CODa \cite{coda} demonstrate that our method surpasses several representative baselines in both accuracy and speed. 
Qualitative results are visualized in Fig. \ref{finalresult}. 
And on the challenging nuScenes detection benchmark \cite{nuscenes}, our detection model UniTR achieves competitive performance among LiDAR-camera fusion methods. 
Furthermore, with minimal additional data collection (only 861 frames) for fine-tuning, our method was successfully transferred to a resource-constrained mobile robot platform, achieving real-time inference while maintaining good accuracy.
\end{itemize}

\begin{figure*}
        \centering
        \includegraphics[width=7in]{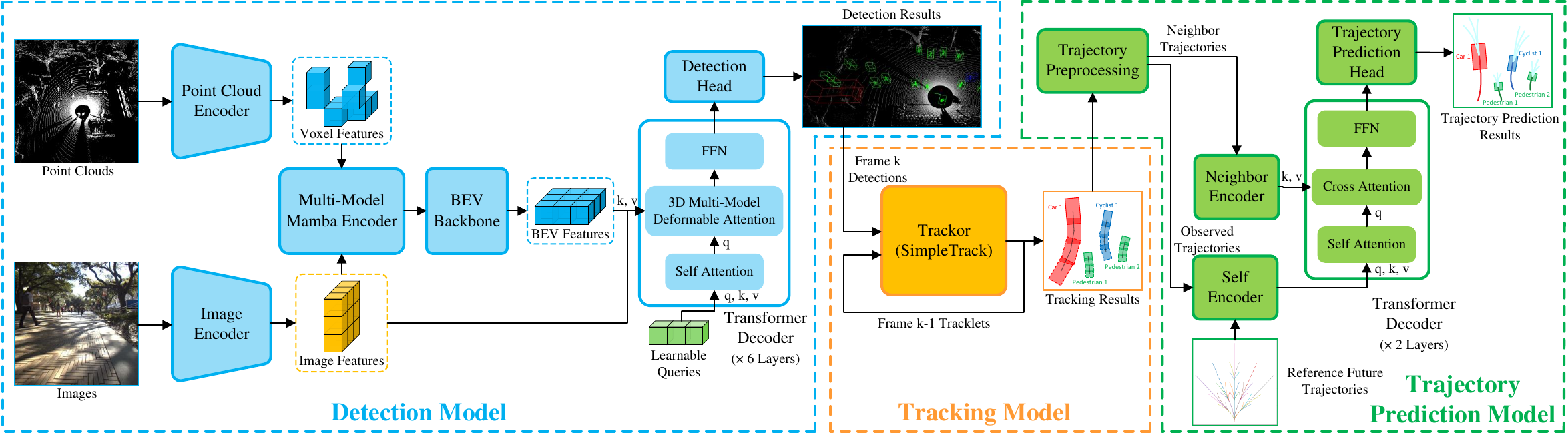}
        \caption{The overall framework of our method. It consists of three main components: the LiDAR-camera fused 3D object detection model UniMT, the object tracking model SimpleTrack, and the multi-class trajectory prediction model RTMCT.}
        \label{wholemodel}
\end{figure*}

\section{RELATED WORK}

\subsection{3D Object Detection}

Early 3D object detection methods \cite{voxelnet,pointpillars,centerpoint,fgf3d} primarily rely on LiDAR as the sensor. 
Following the introduction of Lift-Splat-Shot (LSS) \cite{lss}, camera-based methods gained popularity. 
Both monocular \cite{caddn} and multi-view \cite{bevformer,bevdepth} approaches have demonstrated significant improvements in detection accuracy. 
Additionally, the advent of DETR \cite{detr} spurs the development of methods employing DETR-like decoders, evolving from 2D detection \cite{deformabledetr,dndetr,dino} to camera-based 3D detection \cite{detr3d,petr,monocdit}.

Compared to using a single-sensor method, combining the advantages of LiDAR in geometric information and the camera in semantic information can achieve higher accuracy. 
Early attempts \cite{mv3d,avod,mvxnet} yielded limited performance gains, primarily due to constraints in network architecture and fusion strategies. 

Subsequent research focused on developing more efficient fusion techniques, establishing a clear accuracy advantage for LiDAR-camera fusion models over LiDAR-only counterparts. 
Some point-level methods \cite{pointpainting,virconv} first convert image pixels to 3D virtual points using depth estimation models, then process these virtual points together with LiDAR points. 
The speed and accuracy of the models are significantly constrained by the depth estimation model. 
Some feature-level methods \cite{bevfusion} first transform the image features by LSS, then fuse 3D image features and point cloud features through rigid operations like concatenation or addition. 
The LSS is time-consuming, and the rigid operations of feature fusion are susceptible to the precision of space transformation. 
Based on PETR \cite{petr}, CMT \cite{cmt} extends the DETR-like decoder to multi-modal network, which enables queries to interact with all image and point cloud features through global attention. 
The global attention struggles with large-scale features and may lack precision in capturing fine-grained local features. 
Furthermore, some methods \cite{mv2dfusion,sparselif,isfusion,uafusion} achieve higher accuracy at the cost of highly complex fusion strategies and network architectures, which contradicts the requirements for lightweight design on resource-constrained platforms.

Based on the above discussion, in our detection model UniMT, we introduce both the MME in the backbone and the MDA in the DETR-like decoder. 
Compared to the feature-level methods described above, MME utilizes a more efficient transformation of multi-modal features and avoids rigid fusion operations. 
Compared to the global attention of CMT, MDA can extract more precise features from larger-scale multi-modal features through its sparse attention mechanism. 
At the same time, we introduce the Mamba module \cite{mamba}, which is built upon State Space Models (SSMs) and designed for efficient 1D sequence modeling. 
Its application has been extended to images \cite{visionmamba,vmamba} by serializing 2D feature maps, demonstrating representation power comparable to Transformers while maintaining linear computational complexity. 
Recently, several methods \cite{voxelmamba,lion} employ Mamba to process 3D sparse features in 3D object detection, achieving state-of-the-art performance on LIDAR-only models. 
Inspired by these advances, we leverage Mamba in our MME for efficient multi-modal feature encoding.

\subsection{Object Tracking}
Object tracking methods can be broadly categorized into three frameworks based on their architecture: 1) the Tracking by Detection (TBD) framework; 2) the Joint Detection and Tracking (JDT) framework; 3) the Tracking by Attention (TBA) framework. 
The TBD framework mainly uses the detection results from other methods as input for tracking. 
SORT \cite{sort} tracks objects in 2D images by Kalman filtering and Hungarian matching algorithm. 
AB3DMOT \cite{ab3dmot} extends the TBD tracking framework to 3D scenes, and then improved methods \cite{simpletrack,3dmotformer,camomot} have been proposed. 
For the JDT framework, some methods \cite{centertrack,chainedtracker} get tracking-related information while performing the detection. 
Based on Deformable DETR \cite{deformabledetr}, TransTrack \cite{transtrack} extends the query-based object detection method to an end-to-end detection and tracking method in 2D scenes, creating the Tracking-by-Attention (TBA) framework. 
Further works use this framework for 2D scenes \cite{trackformer,transcenter} and 3D scenes \cite{mutr3d,adatrack}.

Although JDT and TBA methods can leverage richer contextual information, they often entail complex models that are difficult to train and generalize across diverse scenarios. 
In contrast, TBD methods are typically training-free, exhibit fast inference speeds, maintain competitive accuracy, and are easily adaptable to new environments through parameter tuning. 
Therefore, we adopt the TBD method SimpleTrack \cite{simpletrack} as our tracking model.

\subsection{Trajectory Prediction}
Trajectory prediction research initially focused primarily on pedestrian trajectories. Social-LSTM \cite{sociallstm} is the first to achieve reliable trajectory prediction of pedestrians with LSTM. 
Subsequent methods explore alternative architectures; for instance, STGAT \cite{stgat} incorporates graph neural networks with graph attention, while STAR \cite{star} employs dedicated spatial and temporal transformers. 
However, a common limitation of these deterministic models is their inability to generate diverse trajectory predictions. 
In order to incorporate the uncertainty of pedestrians' motions, generative models are introduced. 
Methods like \cite{desire,socialvae,socialimplicit} utilize Conditional Variational Auto-Encoders (CVAEs), while others \cite{socialgan,sophie,iafgan} employ Generative Adversarial networks (GANs). 
However, these generative approaches often suffer from complex architectures, slower inference speeds, and challenging training processes. 
Based on the issues identified in the above methods, we design the trajectory prediction model RTMCT, which can generate diverse future trajectories based on learnable reference trajectories without generative models, and enable fast parallel processing through simple Transformers.
We further address the limitations of prior works by incorporating encoding strategies capable of handling flexible-length histories and multiple object classes.

Beyond pedestrian trajectory prediction, many methods \cite{vectornet,lanercnn,hybridkrcnn} are designed for autonomous driving scenes. 
These methods typically rely on high-definition (HD) map priors and modeling of roads and lane lines to predict vehicle trajectories. 
For mobile service robots operating in daily scenarios, they may not be very suitable.

\section{System Overview}

As illustrated in Fig. \ref{wholemodel}, the proposed framework comprises a detection model, a tracking model, and a trajectory prediction model.
Our novel detection model, named Unified modality detector with Mamba and Transformer (UniMT), takes point clouds from LiDAR and images from monocular or multi-view cameras as input, and produces 3D bounding boxes for pedestrians, vehicles, and riders within the perception range.
For the tracking model, we employ SimpleTrack.
It associates the tracklets from previous frames with the current frame's detection results to generate updated tracklets for all tracked objects.
While the original SimpleTrack runs on the CPU, we accelerate it by implementing its core computational components in parallel on the GPU to meet real-time requirements. 
Our novel trajectory prediction model, named Reference Trajectory-based Multi-Class Transformer (RTMCT), takes the historical trajectories of multiple dynamic objects from the tracking model as input, and generates various possible future trajectories for these objects as the system's final output.
The detection model is detailed in section \ref{ch_detection}, and the trajectory prediction model is elaborated in section \ref{ch_traj_pred}.

\begin{figure*}
        \centering
        \includegraphics[width=7in]{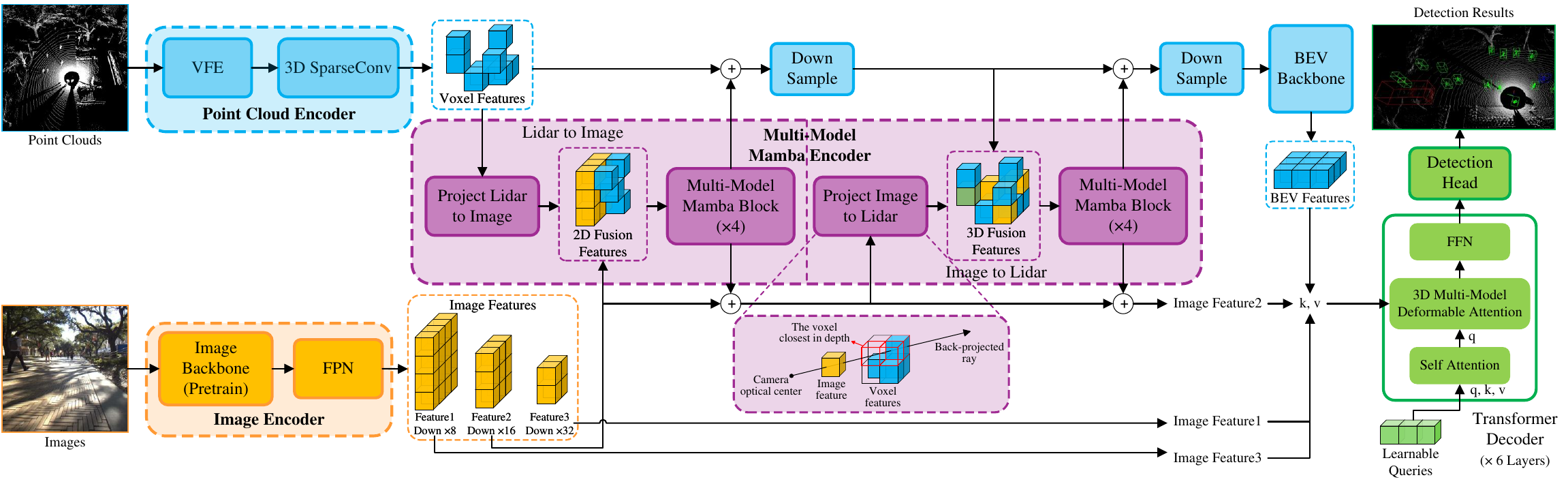}
        \caption{The architecture of our detection model UniMT. It mainly consists of the point cloud encoder, the image encoder, the Multi-model Mamba Encoder (MME), and the DETR-like decoder with 3D Multi-model Deformable Attention (MDA).}
        \label{detectionmodel}
\end{figure*}

\section{Detection} \label{ch_detection}

\subsection{Detection Model Architecture}

The detailed structure of our detection model UniMT is illustrated in Fig. \ref{detectionmodel}. 
The model processes LiDAR point clouds and camera images through dedicated encoders followed by a novel fusion and decoding architecture.

Point clouds are input into a point cloud encoder, which generates 3D sparse voxel features through a voxel feature encoder (VFE, comprising voxelization and PointNet) and a multi-layer 3D sparse convolutional network. 
Images from monocular or multi-view cameras are input into an image encoder, which utilizes a pre-trained backbone and a feature pyramid network (FPN) to produce multi-scale feature maps for each image. 
Specifically, we extract the last three feature maps (at 8×, 16×, and 32× downsampling ratios) from the backbone, and perform multi-scale fusion via FPN. 
The 3D sparse voxel features and the 16× downsampled image features are then fed into the Multi-model Mamba Encoder (MME), which performs feature-level multimodal fusion to produce updated voxel and image features. 
The MME module is detailed in section \ref{ch_mme}.

The updated 3D sparse voxel features are subsequently converted into dense 2D Bird's-Eye View (BEV) features via a BEV backbone (sharing the architecture of the BEV backbone in VoxelNet \cite{voxelnet}). 
Finally, the detection results are generated by a DETR-like decoder, which consists of 6 Transformer layers and a detection head. 
Each Transformer layer comprises a self attention, a 3D Multi-model Deformable Attention (MDA), and a feed-forward network (FFN). 
Following \cite{petr, cmt}, the initial query features and their corresponding 3D reference points are learnable parameters input to the first layer. 
Within each Transformer layer, the queries interact with each other via self attention, and engage with the image and point cloud features through the MDA module. 
The MDA module is detailed in section \ref{ch_mda}.

\subsection{Multi-model Mamba Encoder (MME)} \label{ch_mme}

\begin{figure}
        \centering
        \includegraphics[width=3.4in]{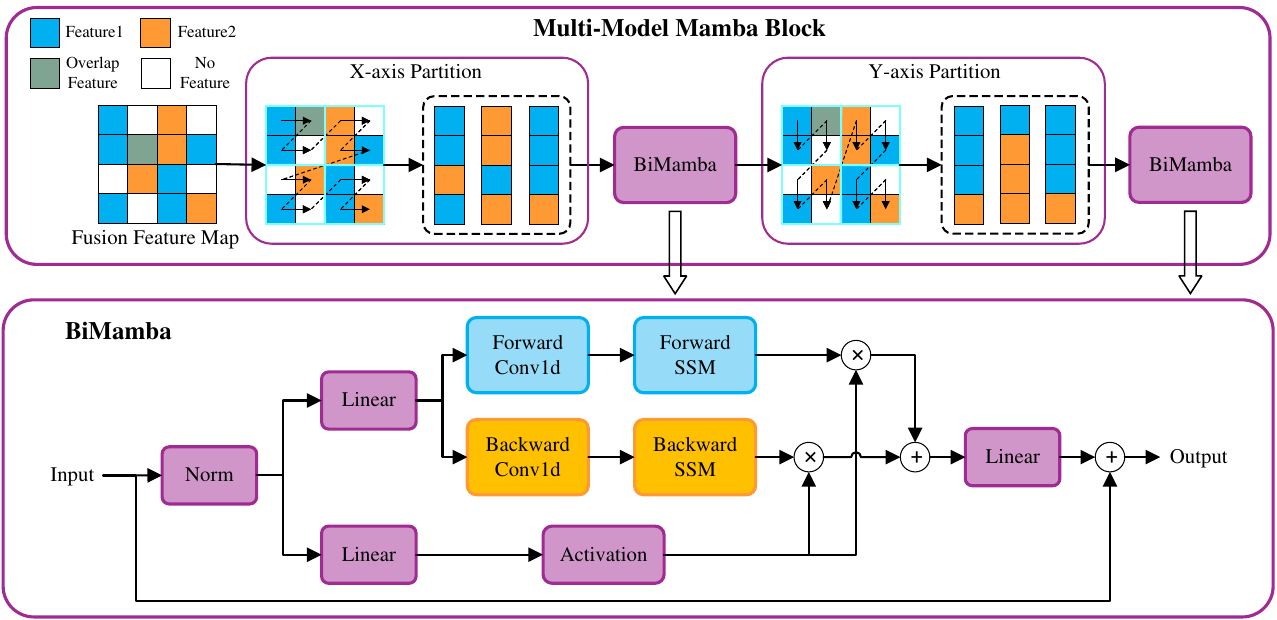}
        \caption{The structure of the Multi-model Mamba Block (MM-Block) used in MME. It serializes and groups the 3D sparse fusion features by X-axis or Y-axis partitions, then encodes them using Bidirectional Mamba (BiMamba). The BiMamba extends the standard Mamba by processing the sequence in both original and reversed orders.}
        \label{mambablock}
\end{figure}

Our novel MME comprises two complementary branches: the ``LiDAR to Image'' branch and the ``Image to LiDAR'' branch. 
In each branch, the MME first projects multi-modal features into unified spaces, then converts these aligned features into 1D sequences with grouping, and finally encodes them through Mamba. 
The Serialization, Grouping, and encoding are performed by a stack of Multi-model Mamba Blocks (MM-Blocks), as illustrated in Fig. \ref{mambablock}.
To maintain a favorable computational overhead, the MME utilizes only the middle-scale feature map (typically at 16× downsampling) from image features. 
The overall process is structured into three key steps, detailed below.

\subsubsection{\textbf{Projections of Multi-modal Features}}
In the ``LiDAR to Image'' branch, the 3D points corresponding to the voxel features are projected onto the 2D image plane using the LiDAR-to-image transformation matrix. 
Only voxel features whose projected coordinates fall within the image boundaries are retained. 
These valid voxel features with 2D coordinates are concatenated with the image features to generate 2D fusion features.

In the ``Image to LiDAR'' branch, image features are projected into 3D space by associating them with the nearest suitable 3D voxels. This is achieved by:
\begin{itemize}
\item Taking the 3D coordinates of all non-empty voxel features and their immediate empty neighbors, and projecting them to 2D image coordinates.
\item For each image feature, identifying the three closest projected 3D candidate points in the 2D image space within a specified radius. 
(Image features with no such points within the radius are excluded from subsequent processing.)
\item Selecting the voxel with the smallest depth value (closest to the camera) from the candidates and assigning its 3D coordinate to the image feature.
\end{itemize}

This process effectively finds the front-most 3D voxel along the back-projected ray of an image feature, with empty neighbors expanding the candidate pool for successful matching. 
These image features with assigned 3D coordinates are then concatenated with the voxel features to produce 3D fusion features.

\subsubsection{\textbf{Serialization and Grouping}}
The fusion features from both branches are uniformly treated as a 3D sparse feature map (treating 2D features in image space with a height of zero along the z-axis), defined by a feature sequence $F \in \mathbb{R}^{N \times D}$ and its 3D coordinate sequence $C \in \mathbb{R}^{N \times 3}$.

As illustrated in Fig. \ref{mambablock}, the MM-Block employs a window-based serialization strategy to convert these sparse features into 1D sequences. 
For serialization by the X-axis partition, the priority order is $x \rightarrow y \rightarrow z$. 
A 3D window of size $[W_x, W_y, W_z]$ is defined. 
For a feature $F_i$ with coordinate $C_i = [x_i, y_i, z_i]$, its corresponding window coordinate is computed by Eq. \ref{eq_mme1}, the unique window index is computed by Eq. \ref{eq_mme2}, and the index within that window is computed by Eq. \ref{eq_mme3}. 

\begin{equation}
        \begin{cases}
        x^{Win}_i = \lfloor x_i / W_x \rfloor, \\
        y^{Win}_i = \lfloor y_i / W_y \rfloor, \\
        z^{Win}_i = \lfloor z_i / W_z \rfloor.
        \end{cases}
        \label{eq_mme1}
\end{equation}

\begin{equation}
        Index^{Win}_i = x^{Win}_i \cdot W_y \cdot W_z + y^{Win}_i \cdot W_z + z^{Win}_i
        \label{eq_mme2}
\end{equation}

\begin{equation}
        \begin{aligned}
        Index^{InWin}_i = (x_i \bmod W_x) \cdot W_y \cdot W_z + \\
        (y_i \bmod W_y) \cdot W_z + (z_i \bmod W_z)
        \end{aligned}
        \label{eq_mme3}
\end{equation}

We serialize features in the order of the window index first, then index within the window. 
This is equivalent to assigning a global sequence index to each feature via Eq. \ref{eq_mme4}. 
Features are sorted by this index to complete serialization. (some overlapping features may be arranged in arbitrary order). 
For the Y-axis partition, the priority order is $y \rightarrow x \rightarrow z$, achieved by swapping $x$ and $y$ in the calculations. 
To preserve the original position information, the serialized features are added with the position encoding of $[x, y, z]$.

\begin{equation}
        Index_i = Index^{Win}_i \cdot (W_x \cdot W_y \cdot W_z) + Index^{InWin}_i
        \label{eq_mme4}
\end{equation}

After serialization, the sequence is padded and divided into $K$ groups of length $group_{size}$, resulting in grouped features $F^g \in \mathbb{R}^{K \times group_size \times D}$ for processing by Mamba.
% Following serialization, the $group\_size$ is defined as the length of each grouped sequence.
% The serialized features are padded to a length $N' = K \times group\_size$, and are divided into $K$ groups.
% The grouped features $F^g \in \mathbb{R}^{K \times group\_size \times D}$ are then processed by Mamba.

\subsubsection{\textbf{Bidirectional Mamba Encoding}}
Given that the serialized features do not exhibit a causal relationship, we employ the BiMamba to capture context from both directions effectively. 
As illustrated in Fig. \ref{mambablock}, the BiMamba architecture extends the standard Mamba by incorporating two separate, non-parameter-sharing 1D convolutional layers and State Space Models (SSMs) to process the sequence in its original and reversed orders. 
The outputs from both directions are then combined. 
The linear computational complexity of Mamba allows $group_size$ to be set to a large value, granting the model a very large effective receptive field compared to CNN or Transformer-based encoders.

Therefore, the MME stacks several MM-Blocks, each alternating between X-axis and Y-axis partition serialization and employing BiMamba for encoding. Consequently, the Multi-modal Mamba Encoder (MME) provides distinct advantages over existing fusion paradigms:
\begin{itemize}
\item Dual-Space Fusion: It performs fusion in both image space (rich in semantics) and 3D LiDAR space (rich in geometry), unlike single-space methods.
\item Efficiency: It bypasses the computational overhead of dense depth estimation and 3D dense feature transformation required by LSS-based methods, resulting in faster processing.
\item Robustness: Its soft fusion via serialization and grouping is less sensitive to sensor miscalibration than rigid operations like concatenation or addition and preserves the native structure of each modality's features.
\item Powerful Context Encoding: The Mamba backbone enables efficient, long-range context modeling over large feature sets with linear complexity, surpassing the limitations of convolutional and Transformer-based encoders.
\end{itemize}

\subsection{3D Multi-model Deformable Attention (MDA)} \label{ch_mda}

\begin{figure}
        \centering
        \includegraphics[width=3.4in]{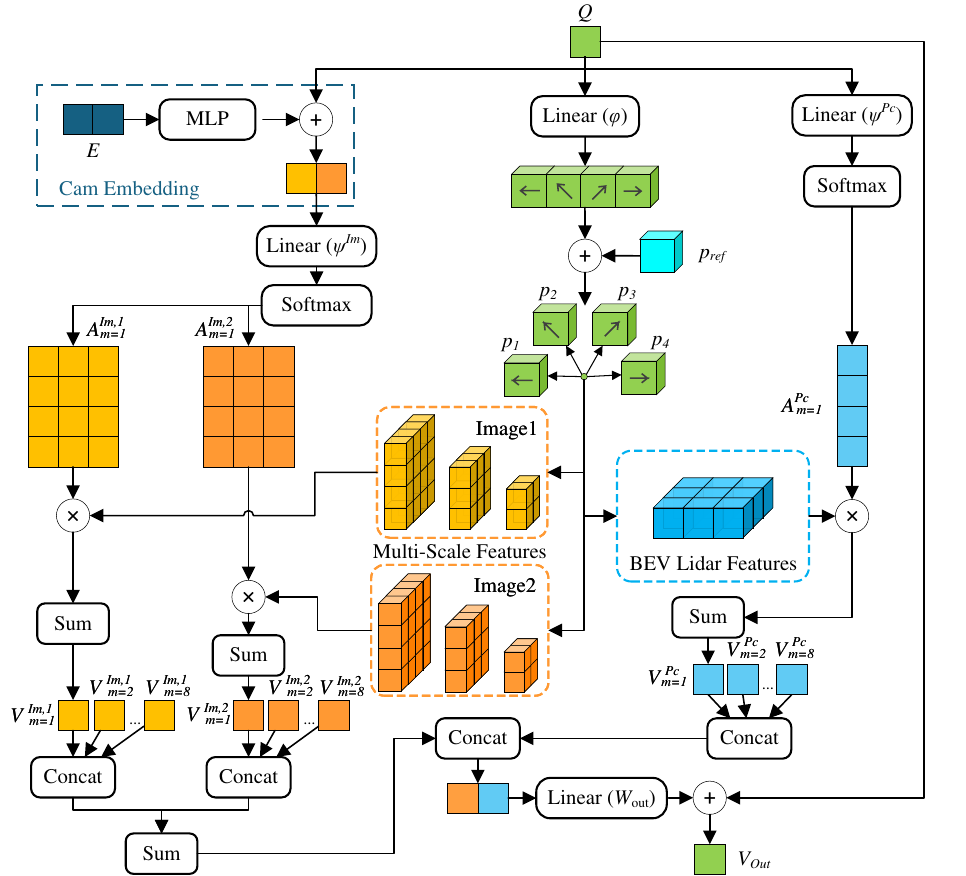}
        \caption{The structure of the 3D Multi-model Deformable Attention (MDA) used in the DETR-like decoder. It extracts and fuses features from the BEV LiDAR feature map and multi-view, multi-scale image feature maps based on 3D sampling points generated from object queries.}
        \label{MDA}
\end{figure}

Building upon Deformable Attention \cite{deformabledetr}, we design the MDA module. Its structure is illustrated in Fig. \ref{MDA}. 
The MDA inputs the object queries and multi-modal features, which include BEV LiDAR features and multi-view, multi-scale image features (the fused 16× feature map and the original 8× and 32× maps). 
In the MDA, each object query generates a set of 3D sampling points based on its 3D reference points and learned offsets. 
It then extracts multi-modal features from these sampled locations and fuses them to update itself.

Formally, let $M$ be the number of attention heads, $J$ be the number of cameras, $L$ be the number of image scales, and $K$ be the number of sampling points. 
In our implementation, we set $M=8$, $J=2$, $L=3$, and $K=8$.
Given an input query $Q$ and its 3D reference point $p$, the $k$-th 3D sampling point $p_{k}$ is computed by Eq. \ref{eq_mda1}, where $\varphi$ is a linear projection layer that outputs the 3D offsets.

\begin{equation}
        p_{k}=\varphi_{k}(Q)+p
        \label{eq_mda1}
\end{equation}

Each sampling point is assigned an attention weight on each feature map. 
The weights for the BEV LiDAR feature map are obtained directly from $Q$ via a linear layer followed by softmax. 
For the image feature maps, the weights are additionally conditioned on the camera parameters to distinguish different views. 
Let $E_{j}$ be denote the intrinsic and extrinsic parameters of the $j$-th camera. 
The weight for the sampling point from the $m$-th head on the $j$-th image is computed by Eq. \ref{eq_mda2}, where $\psi$ is a linear layer followed by softmax.

\begin{equation}
        A_{m}^{I m, j}=\psi_{m}^{I m, j}\left[Q+M L P\left(E_{j}\right)\right]
        \label{eq_mda2}
\end{equation}

Let $p_{k}^{Pc}$ and $p_{k}^{Im,j}$ be the projections of $p_k$ onto the BEV plane and the $j$-th image plane, respectively. 
Let $Pc(x)$ and $Im_{j,l}(x)$ denote the feature vectors at location $x$ obtained via bilinear interpolation on the BEV feature map and the feature map of the $j$-th image at the $l$-th scale, respectively. 
The aggregated feature from the point cloud and the $j$-th image are computed by Eq. \ref{eq_mda3} and Eq. \ref{eq_mda4}.

\begin{equation}
        V_{Pc} = Concat \left\{ \sum_{k=1}^{K} A_{m,k}^{Pc} \cdot Pc(p_{k}^{Pc}); m=1,...,M \right\}
        \label{eq_mda3}
\end{equation}

\begin{equation}
        \footnotesize
        V_{Im,j} = Concat \left\{ \sum_{l=1}^{L} \sum_{k=1}^{K} A_{m,k,l}^{Im,j} \cdot Im_{j,l}(p_{k}^{Im,j}); m=1,...,M \right\}
        \label{eq_mda4}
\end{equation}

Finally, the image features from all views are summed, concatenated with the point cloud feature, and projected to produce the output, as in Eq. \ref{eq_mda5}.

\begin{equation}
        V_{out} = W_{out} \cdot Concat\left(\sum_{j=1}^{J} V_{Im,j}, V_{Pc}\right)
        \label{eq_mda5}
\end{equation}

In the MDA, queries adaptively sample informative features from multi-modal inputs based on potential object properties. 
Its linear complexity relative to the number of sampling points makes it significantly more efficient than global attention while achieving higher precision.

\subsection{Training and Losses}

Following common practices in DETR-like decoders \cite{detr3d,petr,cmt}, the outputs from all 6 Transformer layers in UniMT are passed through the detection head to compute the loss during training. 
The loss function consists of a focal loss for classification and an L1 loss for bounding box regression. 
We employ the denoising strategy \cite{dndetr,dino} to accelerate convergence. 
The pre-trained image backbone is either kept frozen or fine-tuned with a small learning rate.
During inference, only the output from the final layer is used.

\section{Trajectory Prediction} \label{ch_traj_pred}

\begin{figure}
        \centering
        \includegraphics[width=3.4in]{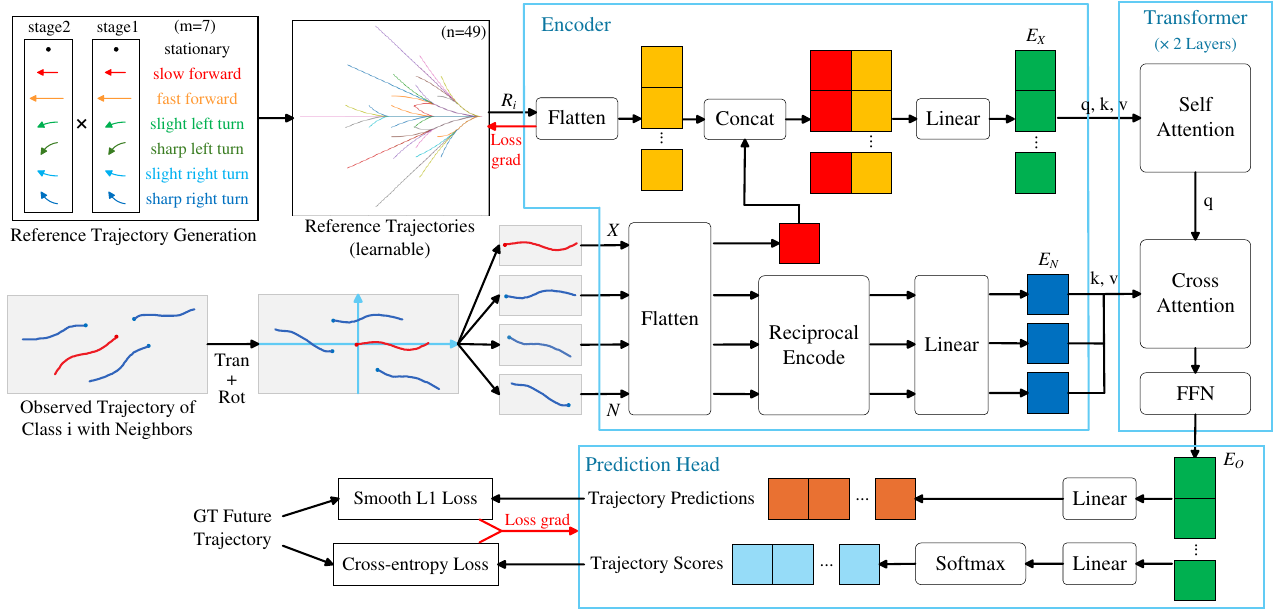}
        \caption{The architecture of our trajectory prediction model RTMCT. It mainly consists of trajectory preprocessing, reference trajectory generation, trajectory encoding, Transformer decoding, and prediction head.}
        \label{trajpredmodel}
\end{figure}

The architecture of our trajectory prediction model RTMCT is shown in Fig. \ref{trajpredmodel}. 
RTMCT takes BEV trajectories of multi-class objects and the robot itself with flexible trajectory lengths as input, and outputs predicted trajectories for these objects. 
Instead of using generative models such as CVAE or GANs, we employ only linear layers and Transformer to encode and decode trajectories in parallel while considering interactions among different objects. 
Diverse trajectory predictions are efficiently generated through reference trajectories. 
We present the model in several parts: trajectory preprocessing, reference trajectory generation, trajectory encoding, Transformer decoding, prediction head and losses.

\subsection{Trajectory Preprocessing}

The model processes each observed trajectory along with its neighbors. 
Let $T_{obs}$ be the length of the input trajectory, and $T_{min}$ be the minimum length of an observed trajectory. 
Trajectories shorter than $T_{min}$ are padded with their earliest position to reach $T_{obs}$. 
Neighbors are selected based on distance thresholds $d_{1},...,d_{c+1}$ defined for each class (including the robot as an additional class). 
For an observed trajectory of class $i$, any trajectory of class $j$ with minimum distance less than $max(d_{i}, d_{j})$ is included as a neighbor. 
Missing positions within $T_{obs}$ are filled with extremely large values.
We thus obtain an observed trajectory $X \in \mathbb{R}^{T_{obs} \times 2}$ and its neighbors $N \in \mathbb{R}^{k \times T_{obs} \times 2}$, where $k$ is the number of neighbors. 
To facilitate model processing, we apply translation and rotation to the observed trajectories and their neighbors such that the last position of the observed trajectory is located at the origin and its earliest position lies on the positive X-axis. 
This ensures that all input observed trajectories have similar positions and orientations.

\subsection{Reference Trajectory Generation}

Reference trajectories serve as the basis for generating diverse predictions, with each representing a potential motion mode. 
For each object, RTMCT predicts its future trajectory within each motion mode according to actual conditions.

Let $T_{pred}$ be the length of the predicted trajectory and $n$ be the number of reference trajectories. 
A set of reference trajectories $R \in \mathbb{R}^{c \times n \times T_{pred} \times 2}$ is generated before training, with their positions as learnable parameters. 
Here, $n$ also denotes the number of predicted future trajectories for each observed trajectory. 
To generate the reference trajectories, we define $m$ distinct motion modes, each characterized by a linear velocity and an angular velocity. 
Each reference trajectory is equally divided into two stages, with each stage adopting one of the $m$ motion modes, resulting in $n = m^2$ reference trajectories in total.

In practice, we set $m=7$, corresponding to 7 motion modes: stationary, slow forward, fast forward, slight left turn, sharp left turn, slight right turn, and sharp right turn. 
This configuration yields $n=49$ reference trajectories. 
The linear velocity of the reference trajectories for different object classes is scaled by a factor based on the motion capability of each class. 
Although the initial distribution of the reference trajectories is manually designed, their learnable nature allows them to adapt to a distribution more suitable for the application scenario, provided that sufficient diversity is ensured at initialization.

\subsection{Trajectory Encoding}

First, the observed trajectory $X$ of class $i$, its neighbors $N$, and the corresponding reference trajectories $R_{i} \in \mathbb{R}^{n \times T_{pred} \times 2}$ are flattened to obtain $X' \in \mathbb{R}^{2T_{obs}}$, $N' \in \mathbb{R}^{k \times 2T_{obs}}$, and $R_{i}' \in \mathbb{R}^{n \times 2T_{pred}}$. 
For the observed trajectory $X'$, it is concatenated with each $R_{i}'$ and passed through a class-specific linear layer $\phi_{i}$ to obtain the encoded self-trajectory $E_{X} \in \mathbb{R}^{n \times d}$, where $d$ is the embedding dimension. 
For the neighbors $N'$, since those closer to the current position of the observed trajectory (i.e., the origin) have greater influence, we take their reciprocal values. 
This reciprocal transformation also converts the extremely large values used for padding in the neighbors into negligible values, preventing abnormal training of the network. 
Each neighbor of class $j$ is encoded through a class-specific linear layer $\psi_{j}$, forming $E_{N} \in \mathbb{R}^{k \times d}$.

\subsection{Transformer Decoding}

The Transformer decoder comprises 2 decoder layers, each consisting of a self attention, a cross attention, and a feed-forward network (FFN).
Both the self attention and the cross attention are regular attentions. 
The self-trajectory $E_{X}$ serves as the initial queries for the transformer decoder. 
In the self attention, queries associated with different reference trajectories interact to evaluate the plausibility of each predicted trajectory. 
In the cross attention, the self-trajectory interacts with neighbors to refine its predictions. 
The output of the Transformer decoder is $E_{O} \in \mathbb{R}^{n \times d}$.

\subsection{Prediction Head and Losses}

The prediction head takes $E_{O}$ as input and consists of two branches: the trajectory prediction head and the score head. 
The trajectory prediction head is a linear layer that outputs future trajectory predictions $Y' \in \mathbb{R}^{n \times T_{pred} \times 2}$, while the score head is a linear layer followed by softmax that outputs confidence scores for each prediction. 
Different linear layers are used for different object classes. 
During training, given a ground-truth future trajectory $Y \in \mathbb{R}^{T_{pred} \times 2}$, the Average Displacement Error (ADE) between $Y$ and each of the $n$ reference trajectories $R_{i}$ is computed. 
The output associated with the reference trajectory having the smallest ADE is selected as the positive sample, as shown in Eq. \ref{eq_traj1}. 
The loss function combines trajectory prediction loss and score loss. 
The trajectory prediction loss is the Smooth L1 loss between the positive sample $Y_{t}'$ and $Y$, and the score loss is the cross-entropy loss for the positive sample with a ground-truth value of 1.

\begin{equation}
        t = \operatorname*{argmin}_{t \in \{ 1, \ldots, n \}} \| Y - R_{i,t} \|_2
        \label{eq_traj1}
\end{equation}

\section{EXPERIMENTS}

In this section, we evaluate the performance of our proposed method on both 3D object detection and trajectory prediction tasks using the campus dataset CODa \cite{coda} and the autonomous driving dataset nuScenes \cite{nuscenes}. 
We further assess its practical deployment potential on a mobile robot platform. 
Additional ablation studies are conducted to verify the effectiveness of the key components within our framework.

\subsection{Public Dataset}

The CODa dataset \cite{coda} is utilized for our experiments on both 3D object detection and trajectory prediction.
It is collected in a campus environment using a mobile robot as the perception platform, encompassing both indoor and outdoor scenes under varying lighting and weather conditions. 
The dataset provides annotations for 53 classes for 3D object detection at a collection rate of 10 Hz. 
For our experiments, we use the data captured by a 128-beam LiDAR and two RGB cameras. 
The dataset is split into training, validation, and testing sets comprising 18,050, 4,906, and 4,907 frames, respectively. 
We focus on perceiving dynamic objects and adopt the same class definitions as the KITTI dataset \cite{kitti}: Pedestrian, Car, and Cyclist. 
Specifically, all vehicle-related classes in CODa are unified as Car, while classes such as Bike and Motorcycle are unified as Cyclist. 
The perception range is set to $[-21.0\text{m}, 21.0\text{m}]$ along the X and Y axes, and $[-2.0\text{m}, 6.0\text{m}]$ along the Z axis. 
The dataset contains 573,697 Pedestrian, 43,363 Car, and 8,355 Cyclist valid labels.

To further validate the effectiveness of our detection model UniMT, We conduct additional experiments on the nuScenes dataset \cite{nuscenes}. 
nuScenes is a large-scale autonomous driving dataset collected in urban environments, featuring 1,000 scenes with 10 classes annotated for 3D object detection. 
The dataset is divided into training, validation, and testing sets containing 700, 150, and 150 scenes, respectively. 
Following the nuScenes detection benchmark, we utilize the 32-beam LiDAR and six surround-view cameras, set the perception range to $[-51.2\text{m}, 51.2\text{m}]$ along the X and Y axes, and $[-5.0\text{m}, 3.0\text{m}]$ along the Z axis, and evaluate the detection performance on all 10 classes.

\subsection{Detection Experiment}

\subsubsection{Detection on CODa}

\begin{table*}[t]
	\centering
	\caption{Performance of detection models on CODa}
	\resizebox{150mm}{10mm}{
        \begin{tabular}{cccccccc}
                \hline
                Methods & Modality & AP(Pedestrian) & AP(Car) & AP(Cyclist) & ~~mAP~~  & \#Params & Inference Time \\ \hline
                CenterPoint\cite{centerpoint} & L & 75.56 & 52.56 & 60.64 & 62.92 & \textbf{7.8 M} & \textbf{36 ms} \\ 
                BEVFusion\cite{bevfusion} & L+C & 79.55 & 54.18 & 63.74 & 65.82 & 39.3 M & 176 ms \\ 
                CMT\cite{cmt} & L+C & 80.87 & 62.41 & 66.38 & 69.89 & 82.5 M & 147 ms \\ 
                UniMT (ours) & L+C & \textbf{81.42} & \textbf{66.25} & \textbf{73.14} & \textbf{73.60} & 25.9 M & 139 ms \\ \hline
        \end{tabular}
	}
	\label{tab_detection}
\end{table*}

\begin{figure}
	\centering
	\includegraphics[width=3.4in]{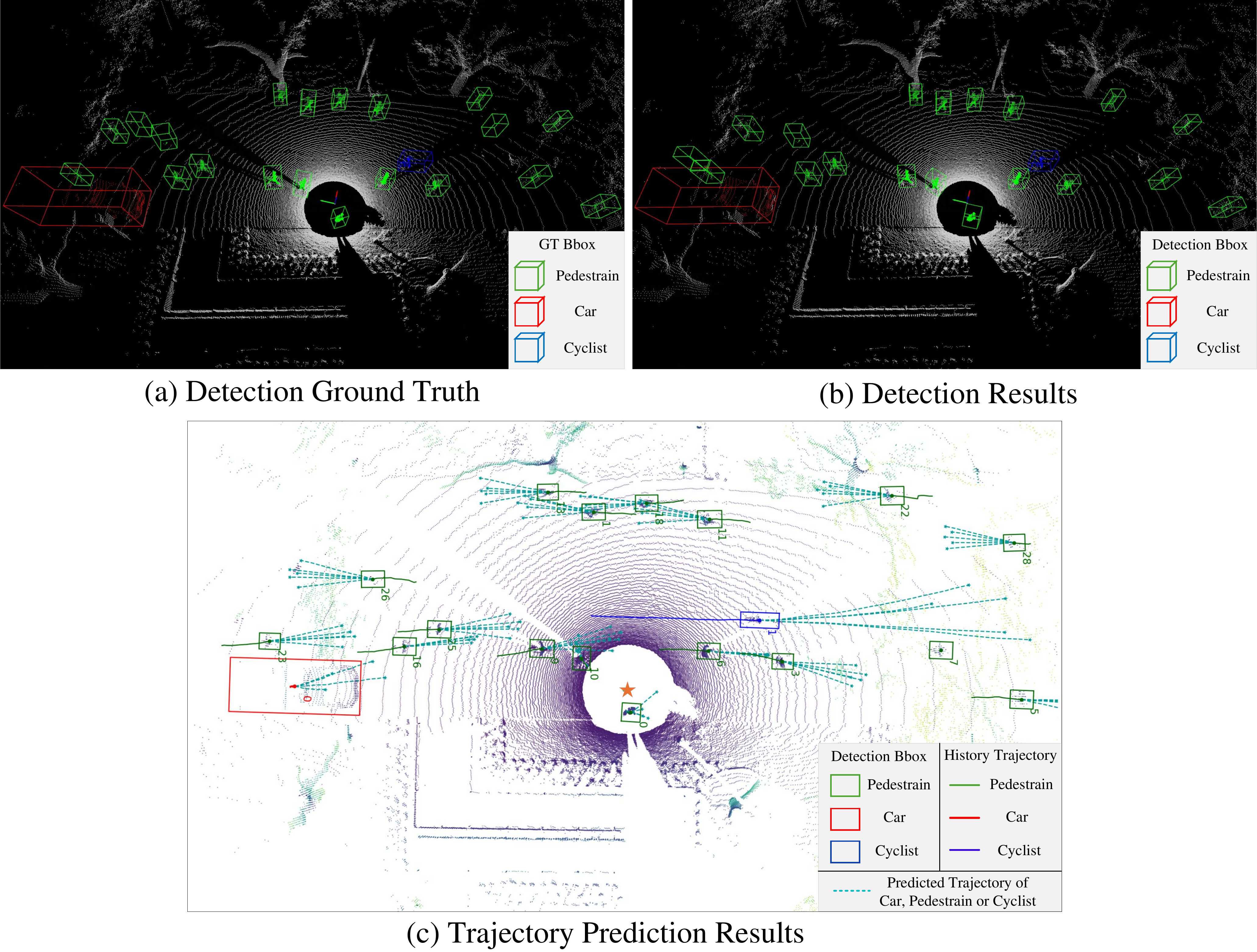}
	\caption{Visualization of detection and trajectory prediction results of our method for the same frame on CODa. (a) labels the 3D Bboxes of ground truth in the point cloud, while (b) labels the 3D Bboxes of detection. (c) shows the object Bboxes, object IDs, history trajectories, and predicted trajectories with top-5 confidence scores of the tracked objects in BEV.}
	\label{codaresult}
\end{figure}

\begin{figure*}
        \centering
        \includegraphics[width=7in]{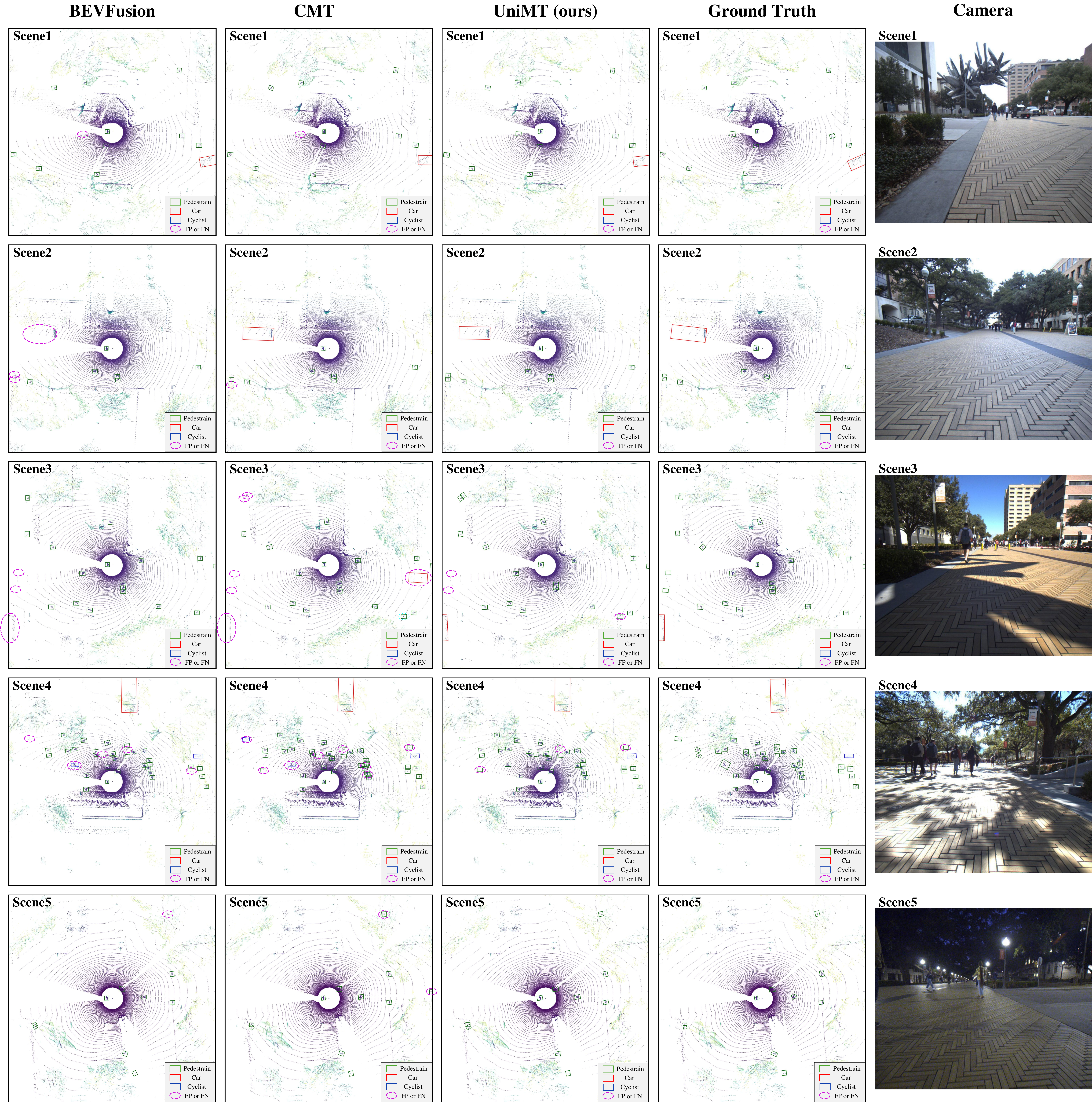}
        \caption{Constrast of detection results between representative baselines and our UniMT on CODa. In Scene1 and Scene2 of cloudy conditions, UniMT successfully detects all objects without false positives, while other methods miss some objects or produce false detections. In Scene3 and Scene4 of sunny conditions, although all methods make some mistakes, UniMT has the least missed detections and false detections. In Scene5 of night conditions, UniMT also accurately detects all objects.}
        \label{detectresults}
\end{figure*}

We train and evaluate our detection model, UniMT, on the CODa dataset. 
UniMT employs a pre-trained ConvNeXtv2-nano \cite{convnextv2} as the image backbone and utilizes 500 object queries in the decoder. 
The model is trained for 30 epochs on 2 NVIDIA GeForce RTX 3090 GPUs using the Class Balanced Group Sampling (CBGS) strategy, the AdamW optimizer, a cosine annealing learning rate scheduler, and an initial learning rate of $3.5 \times 10^{-4}$. 
We compare UniMT with several representative baselines, including the LiDAR-only model CenterPoint \cite{centerpoint} and the multi-modal models BEVFusion \cite{bevfusion} and CMT \cite{cmt}.

The performance of these models on the CODa test set is summarized in Table \ref{tab_detection}. 
We adopt the Bird's-Eye View (BEV) Average Precision (AP) metric, following the KITTI benchmark \cite{kitti}, with IoU thresholds of 0.5, 0.7, and 0.5 for Pedestrian, Car, and Cyclist, respectively. 
The mean Average Precision (mAP) is computed as the average of the three class-wise APs. 
Inference time is measured on an NVIDIA GeForce RTX 3060 GPU.

As a structurally simple LiDAR-only model, CenterPoint has the smallest number of parameters and the fastest inference speed, but at the cost of lower detection accuracy. 
By incorporating image information, BEVFusion improves detection accuracy across all classes compared to CenterPoint, but suffers from a significant increase in inference time. 
Furthermore, due to inaccuracies in the LiDAR-camera transformation matrix within the CODa dataset, the LSS fusion strategy employed by BEVFusion - which is sensitive to such errors - yields only limited accuracy improvements. 
In contrast, CMT achieves higher accuracy than BEVFusion through a more robust implicit fusion strategy. 
Although CMT has a large number of parameters, they are primarily in 2D convolutional networks. So it achieves faster inference than BEVFusion.

Our UniMT outperforms these baselines in detection accuracy by leveraging efficient and deep multi-modal fusion at both the encoder and decoder levels, while also achieving a reduction in model parameters and inference time. 
This advantage is particularly pronounced for the Car class (which requires a high IoU threshold) and the Cyclist class (which has limited training samples). 
These results demonstrate that UniMT is not only easy to train and robust to LiDAR-camera calibration errors, but also achieves higher accuracy with reduced inference time. 
Fig. \ref{codaresult} (a) and (b) visualize the detection results of UniMT in a dynamic scenario, confirming its effectiveness. 
Fig. \ref{detectresults} compares the detection results of UniMT with those of representative baselines in various scenarios, further illustrating its superior performance and robustness for different lighting conditions.

\subsubsection{Detection on nuScenes}

\begin{table*}[t]
	\centering
	\caption{Performance of detection models on the nuScenes test set. \\ All methods don't use test-time augmentation and model ensemble.}
	\resizebox{150mm}{22mm}{
        \begin{tabular}{c|c|ccccc|cc}
                \hline
                Methods & Modality & mATE$\downarrow$ & mASE$\downarrow$ & mAOE$\downarrow$ & mAVE$\downarrow$ & mAAE$\downarrow$ & mAP$\uparrow$ & \textcolor{red}{NDS} $\uparrow$ \\ \hline
                TransFusion\cite{transfusion} & L+C & 25.9 & 24.3 & 35.9 & 28.8 & 12.7 & 68.9 & 71.7 \\ 
				AutoAlignV2\cite{autoalignv2} & L+C & 24.5 & \textbf{23.3} & 31.1 & 25.8 & 13.3 & 68.4 & 72.4 \\
                BEVFusion\cite{bevfusion} & L+C & 26.1 & 23.9 & 32.9 & 26.0 & 13.4 & 70.2 & 72.9 \\ 
				MSMDFusion\cite{msmdfusion} & L+C & 25.5 & 23.8 & 31.0 & 24.4 & 13.2 & 71.5 & 74.0 \\
                CMT\cite{cmt} & L+C & 27.9 & 23.5 & 30.8 & 25.9 & 11.2 & 72.0 & 74.1 \\ 
				EA-LSS\cite{ea-lss} & L+C & 24.7 & 23.7 & 30.4 & 25.0 & 13.3 & 72.2 & 74.4 \\
				UniTR\cite{unitr} & L+C & 24.1 & 22.9 & \textbf{25.6} & 24.0 & 13.1 & 70.9 & 74.5 \\
				FocalFormer3D-F\cite{focalformer3d} & L+C & 25.1 & 24.2 & 32.8 & 22.6 & 12.6 & 72.4 & 74.5 \\
				DAL\cite{dal} & L+C & 25.3 & 23.8 & 33.4 & \textbf{17.4} & 12.0 & 72.0 & 74.8 \\
				FusionFormer\cite{fusionformer} & L+C+T & 26.7 & 23.6 & 28.6 & 22.5 & \textbf{10.5} & 72.6 & 75.1 \\
                UniMT (ours) & L+C & \textbf{23.9} & 23.4 & 27.4 & 23.3 & 12.7 & \textbf{72.7} & \textcolor{red}{75.3} \\ \hline
        \end{tabular}
	}
	\label{tab_nuscenes}
\end{table*}

We further train and evaluate our UniMT model on the nuScenes dataset. 
The experimental settings are adjusted accordingly: we utilize a ConvNeXtv2-tiny \cite{convnextv2} as the image backbone, employ 900 object queries in the decoder, double the layers of the MME from 4 to 8 to handle the more complex scenes, and train the model for 48 epochs without the CBGS strategy. 
The performance of UniMT and several representative state-of-the-art baselines on the nuScenes test set is summarized in Table \ref{tab_nuscenes}. 
The evaluation follows the official nuScenes detection benchmark \cite{nuscenes} metrics, including the nuScenes detection score (NDS) and mean Average Precision (mAP), along with several true positive metrics. 
Our UniMT achieves competitive performance with a mAP of 72.7\% and NDS of 75.3\%, surpassing most LiDAR-camera fusion methods. 
This result demonstrates the strong generalization capability of our proposed fusion strategy and model architecture when applied to different environmental conditions and sensor configurations.

\subsection{Trajectory Prediction Experiment}

\begin{table*}[t]
	\centering
	\caption{Performance of trajectory prediction models on CODa. \\ RTMCT* uses tracking results as input. Other models use ground truth as input.}
	\resizebox{176mm}{7.2mm}{
        \begin{tabular}{ccccccccc}
                \hline
                Methods & Input & ADE(Pedestrian)$_{3/5/10}$ & ADE(Car)$_{3/5/10}$ & ADE(Cyclist)$_{3/5/10}$ & FDE(Pedestrian)$_{3/5/10}$ & FDE(Car)$_{3/5/10}$ & FDE(Cyclist)$_{3/5/10}$ & Inference Time \\ \hline
                Social-GAN\cite{socialgan} & Ground Truth & 0.30/0.28/0.26 & 1.12/1.07/1.01 & 1.67/1.54/1.39 & 0.52/0.48/0.44 & 2.07/1.97/1.85 & 3.22/2.93/2.61 & 52 ms \\ 
                Social-Implicit\cite{socialimplicit} & Ground Truth & 0.35/0.33/0.32 & 0.82/0.80/0.78 & 1.13/1.11/1.09 & 0.62/0.58/0.54 & 1.52/1.49/1.45 & 2.12/2.09/2.04 & 99 ms \\ 
                RTMCT (ours) & Ground Truth & \textbf{0.27/0.24/0.21} & \textbf{0.39/0.35/0.31} & \textbf{0.85/0.78/0.72} & \textbf{0.47/0.41/0.34} & \textbf{0.70/0.61/0.52} & \textbf{1.75/1.59/1.45} & \textbf{35 ms} \\ \hline
                RTMCT* (ours) & Tracking Results & 0.25/0.22/0.19 & 0.42/0.38/0.35 & 0.86/0.77/0.70 & 0.43/0.37/0.31 & 0.68/0.61/0.53 & 1.75/1.57/1.41 & 35 ms \\ \hline
        \end{tabular}
	}
	\label{tab_traj_pred}
\end{table*}

\begin{figure*}
	\centering
	\includegraphics[width=6in]{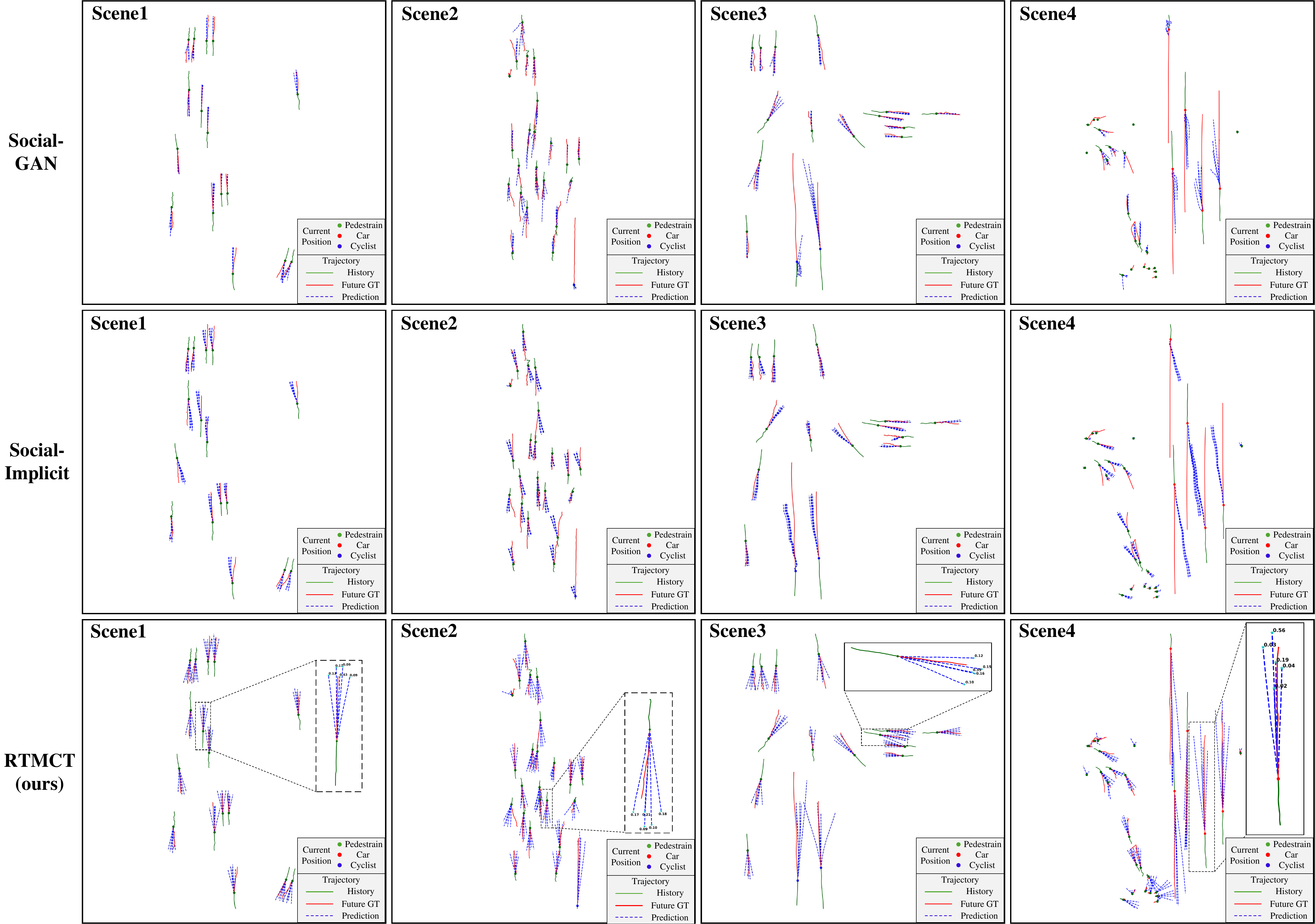}
	\caption{Constrast of trajectory prediction results between representative baselines and our RTMCT on CODa. In scene1 and scene2, compared to Social-GAN and Social-Implicit, RTMCT predicts more diverse and reliable trajectories for many pedestrians. There are cyclists in Scene 3 and cars in Scene 4. RTMCT also predicts the diverse and reliable trajectories for cyclists and cars, while Social-GAN and Social-Implicit perform poorly or even appear abnormal predictions due to the lack of multi-class support and limited training samples. 
			In each scene, we also show a zoomed-in view of a specific object with confidence scores of predictions.}
	\label{trajpredresult}
\end{figure*}

For the trajectory prediction experiment, we compare our RTMCT model with Social-GAN \cite{socialgan} and Social-Implicit \cite{socialimplicit}. 
We use the ground truth trajectories from the CODa training and testing splits to train and evaluate these models. 
For RTMCT, we also employ SimpleTrack to obtain tracking results based on UniMT's detections, and use these tracking results as inputs (with the ground truth as the target) to evaluate its performance in a practical deployment scenario. 
Since Social-GAN and Social-Implicit do not support multi-class input, they are trained and evaluated without distinguishing trajectory classes.

We set the observation length $T_{obs}=16$, the minimum trajectory length $T_{min}=2$, and the prediction horizon $T_{pred}=24$. 
For RTMCT, the neighbor distance thresholds for Pedestrian, Car, Cyclist, and the robot itself are set to 2m, 5m, 3m, and 2m, respectively. 
The number of reference trajectories $n$ is 49, and the Transformer decoder comprises 2 layers. 
The model is trained on an NVIDIA GeForce RTX 3060 GPU using the CBGS strategy and a learning rate of $1.0 \times 10^{-4}$.

The performance of these models on the CODa test set is presented in Table \ref{tab_traj_pred}. 
We report the minimum Average Displacement Error (ADE) and Final Displacement Error (FDE) for the top-3, top-5, and top-10 trajectory predictions. 
Inference time is measured on an RTX 3060 GPU. 
When using tracking results as input, the negative samples do not have matching truth values. 
So we only compute metrics for positive samples. 
Since Social-GAN and Social-Implicit do not distinguish object classes, they perform poorly on Car and Cyclist trajectories, which have a small number of samples. 
In contrast, our RTMCT achieves high trajectory prediction accuracy across all classes while maintaining remarkably fast inference speeds. 
Moreover, when provided with tracking results as input, RTMCT's accuracy remains nearly on par with that achieved using ground truth inputs. 
This indicates that although our model is trained on ground truth trajectories, it generalizes well to practical scenarios with tracked trajectories. 
Fig. \ref{codaresult} (c) visualizes a frame of trajectory prediction results generated by RTMCT. 
Fig. \ref{trajpredresult} compares the trajectory prediction results of RTMCT with those of representative baselines in various scenarios, demonstrating its capability to predict diverse and reliable trajectories for multi-class objects.

\subsection{Experiments on Wheelchair Robot}

\begin{figure}
	\centering
	\includegraphics[width=3in]{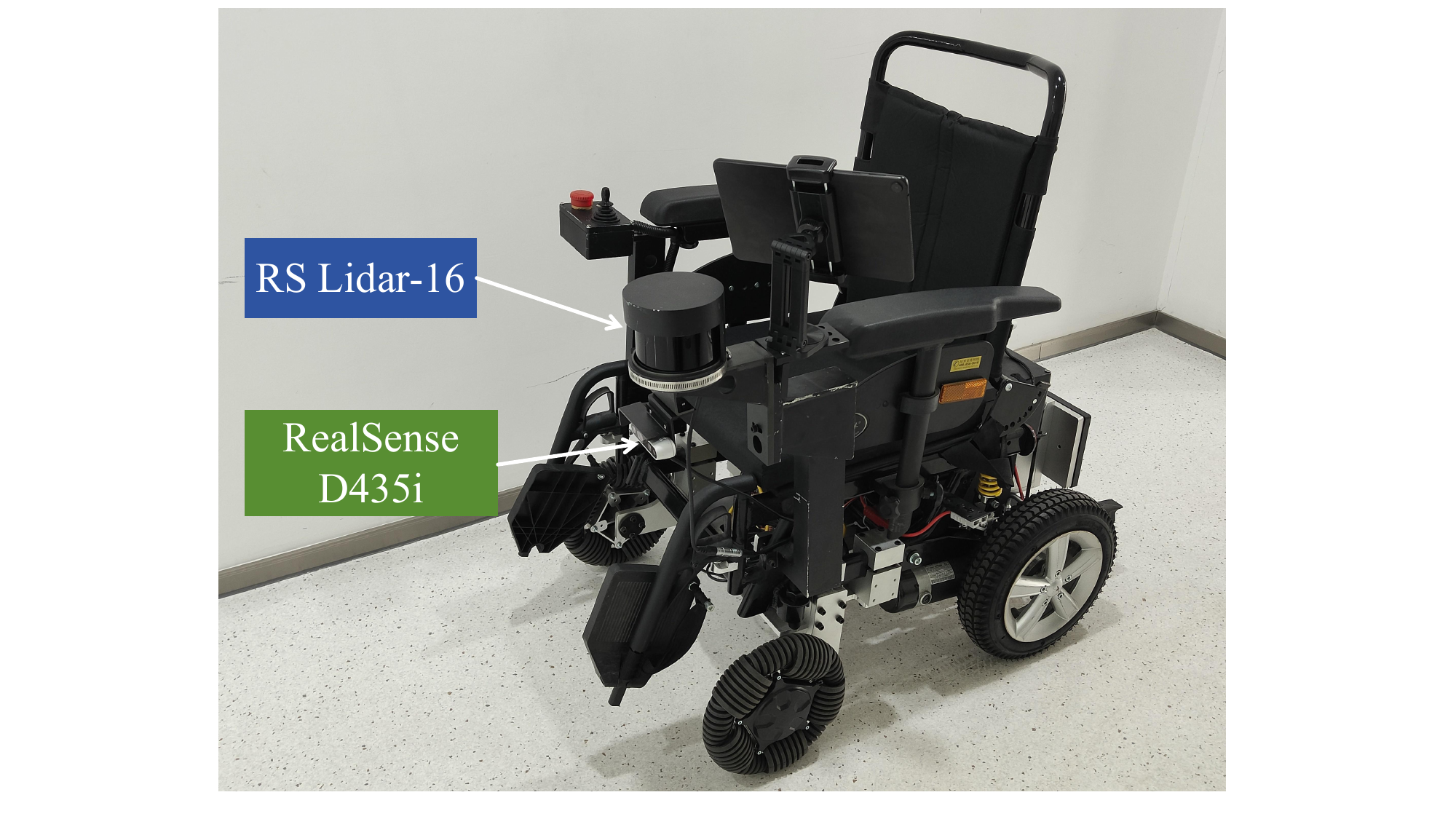}
	\caption{Our intelligent wheelchair robot platform. It is equipped with a 16-beam LiDAR and an RGB-D camera for perception.}
	\label{wheelchair}
\end{figure}

\begin{table}[t]
	\centering
	\caption{Performance of our detection model on wheelchair dataset}
	\resizebox{65mm}{4.2mm}{
        \begin{tabular}{cccc}
                \hline
                Methods & AP(Pedestrian) & AP(Cyclist) & mAP \\ \hline
                UniMT & 78.48 & 75.63 & 77.06 \\ \hline
        \end{tabular}
	}
	\label{tab_detection_wheelchair}
\end{table}

\begin{table}[t]
	\centering
	\caption{Performance of our trajectory prediction model on wheelchair dataset}
	\resizebox{75mm}{9mm}{
        \begin{tabular}{ccc}
                \hline
                Methods & ADE(Pedestrian)$_{3/5/10}$ & ADE(Cyclist)$_{3/5/10}$  \\ \hline
                RTMCT & 0.41/0.36/0.32 & 0.96/0.83/0.74 \\ \hline
                \\ \hline
                        & FDE(Pedestrian)$_{3/5/10}$ & FDE(Cyclist)$_{3/5/10}$ \\ \hline
                RTMCT  & 0.77/0.67/0.59 & 1.83/1.58/1.36 \\ \hline
        \end{tabular}
	}
	\label{tab_traj_pred_wheelchair}
\end{table}

We select an intelligent wheelchair robot as our mobile platform, as shown in Fig. \ref{wheelchair}. 
The robot is equipped with a 16-beam LiDAR (RS Lidar-16), an RGB-D camera (RealSense D435i), an Intel i5-12400F CPU, and an NVIDIA GeForce RTX 3060 GPU.

Compared to the sensors used in CODa, our LiDAR has fewer beams (16 vs. 128) and our camera has a lower resolution ($640 \times 480$ vs. $1224 \times 1024$). 
We set the perception range to $[0.0\text{m}, 15.0\text{m}]$ along the X-axis and $[-10.0\text{m}, 10.0\text{m}]$ along the Y-axis. 
We collect and annotate 1,211 data frames outdoors using the wheelchair robot to form the wheelchair dataset. 
The ego-motion compensation is handled using a LiDAR-based localization method \cite{roll}. 
Due to the scarcity of Car instances in the operating environments, the wheelchair dataset contains only 2,600 Pedestrian labels and 1,302 Cyclist labels. 
The dataset is split into training and testing sets of 861 and 350 frames, respectively.

The wheelchair dataset is too small to train the detection model UniMT from scratch. 
Therefore, we employ a multi-stage fine-tuning strategy. 
We first adapt the CODa dataset to match the wheelchair dataset's specifications by downsampling the point clouds (from 128-beam to 16-beam density) and adjusting the input image resolution and point cloud range accordingly. 
We also reduce the number of decoder queries in UniMT from 500 to 150. 
The model is first pre-trained on the original CODa, then fine-tuned on the adapted CODa dataset, and finally fine-tuned on the wheelchair training set. 
The performance of the resulting UniMT model on the wheelchair test set is reported in Table \ref{tab_detection_wheelchair}, demonstrating detection accuracy comparable to that on the CODa test set (Section IV-B) with the smaller perception range but inferior sensors.

For trajectory prediction, the RTMCT model pre-trained on CODa can be directly applied to the wheelchair robot without further fine-tuning. 
Its performance on the wheelchair test set is shown in Table \ref{tab_traj_pred_wheelchair}. 
The accuracy is slightly inferior to that reported in Section IV-C, which is expected due to the domain shift. 
Nevertheless, it still meets the requirements for practical application.

\begin{figure}
	\centering
	\includegraphics[width=3.4in]{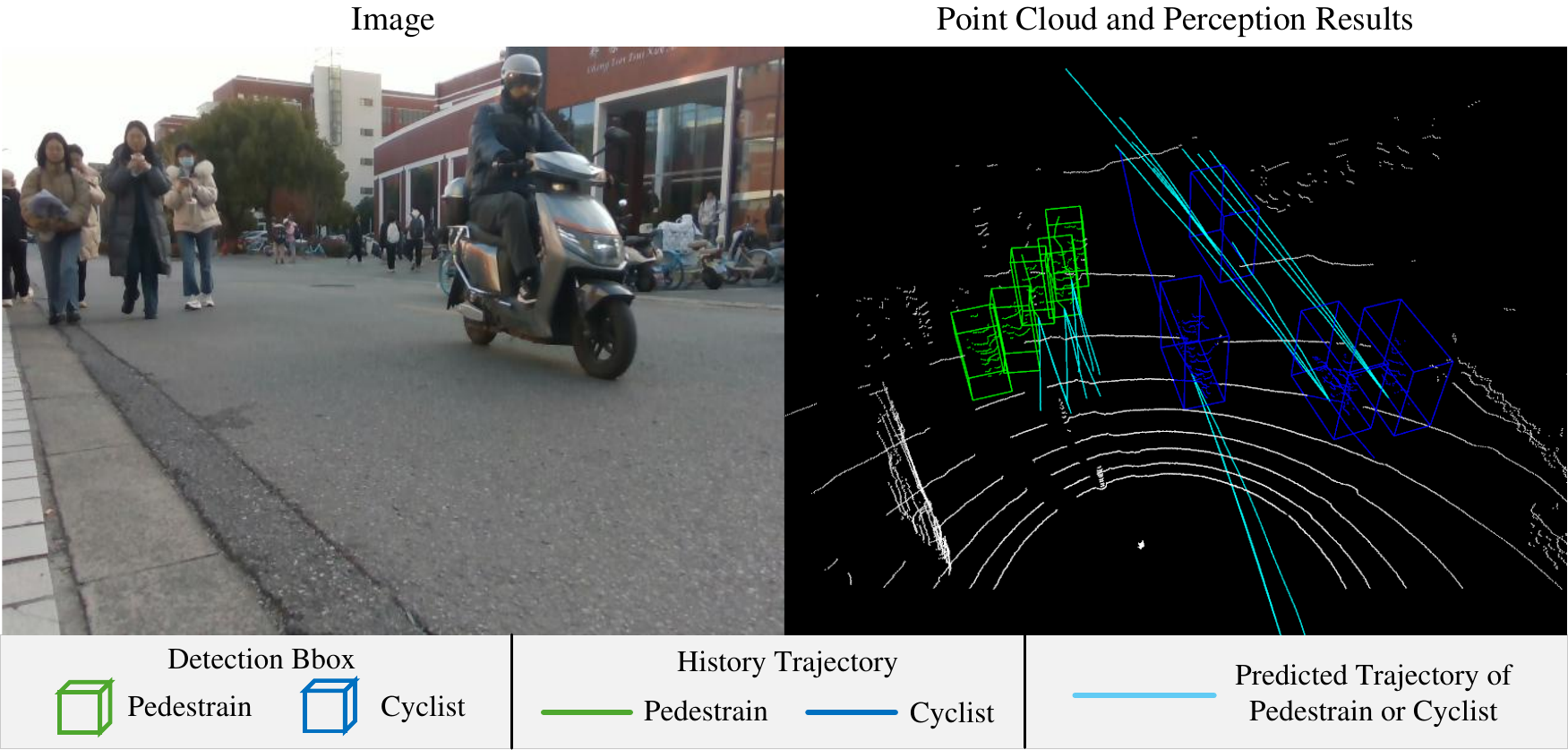}
	\caption{Detection and trajectory prediction results of our method on the wheelchair robot.}
	\label{finalresultwheelchair}
\end{figure}

We deploy the entire framework on the wheelchair robot and conduct real-world tests in a campus environment, as shown in Fig. \ref{finalresultwheelchair}. 
The inference speed of each component is detailed in Table \ref{tab_infer_time_wheelchair}. 
Notably, our GPU re-implementation of the SimpleTrack tracker, which reduces its inference time from 39.9 ms (on CPU) to merely 3.6 ms, achieving an approximately 11× speedup. 
This optimization ensures that the tracking module adds negligible latency to the entire pipeline. 
Consequently, the complete system achieves an overall processing rate of 13.9 frames per second (FPS), which comfortably satisfies real-time requirements. 
Additionally, a supplementary video submitted with this paper showcases the real-time inference results of our method on the wheelchair robot in both indoor and outdoor environments.

\begin{table}[t]
	\centering
	\caption{Inference time of our method on the wheelchair}
	\resizebox{85mm}{3.2mm}{
		\begin{tabular}{ccccc}
			\hline
			Task & Detection & Tracking & Trajectory Prediction & Total \\ \hline
			Inference Time & 64.3 ms & 3.6 ms (39.9 ms on CPU) & 4.3 ms & 72.2 ms \\ \hline
		\end{tabular}
	}
	\label{tab_infer_time_wheelchair}
\end{table}

\subsection{Ablation Study}

To validate the effectiveness of the key components in our proposed models, we conduct comprehensive ablation studies on the CODa dataset for both the detection model and the trajectory prediction model.

\subsubsection{\textbf{Ablation Study for Detection Model}}

\begin{table}[t]
	\centering
	\caption{Ablation study on the MME module in UniMT}
	\resizebox{85mm}{8mm}{
		\begin{tabular}{cccccc}
			\hline
			L2I Branch & I2L Branch & AP(Pedestrian) & AP(Car) & AP(Cyclist) & mAP \\ \hline
			- & - & 80.95 & 64.23 & 70.35 & 71.84 \\
			\checkmark & - & 81.13 & 65.38 & 71.58 & 72.70 \\ 
			- & \checkmark & 81.25 & 65.77 & 72.32 & 73.11 \\
			\checkmark & \checkmark & \textbf{81.42} & \textbf{66.25} & \textbf{73.14} & \textbf{73.60} \\ \hline
		\end{tabular}
	}
	\label{tab_ablation_mme}
\end{table}

\begin{table}[t]
	\centering
	\caption{Ablation study on the MDA module in UniMT}
	\resizebox{85mm}{7.5mm}{
		\begin{tabular}{ccccc}
			\hline
			Fusion Type & AP(Pedestrian) & AP(Car) & AP(Cyclist) & mAP \\ \hline
			Bilinear Sampling & 79.56 & 61.67 & 65.49 & 68.91 \\
			Global Attention &  80.91 & 64.84 & 70.91 & 72.22 \\ 
			MDA (Ours) & \textbf{81.42} & \textbf{66.25} & \textbf{73.14} & \textbf{73.60} \\ \hline
		\end{tabular}
	}
	\label{tab_ablation_mda}
\end{table}

\begin{table}[t]
	\centering
	\caption{Ablation study on image feature processing in UniMT}
	\resizebox{85mm}{6.7mm}{
		\begin{tabular}{ccccccc}
			\hline
			Configuration & Pre-trained & Multi-scale & AP(Pedestrian) & AP(Car) & AP(Cyclist) & mAP \\ \hline
			A & - & - & 80.83 & 63.51 & 69.14 & 71.16 \\
			B & - & \checkmark & 80.91 & 64.37 & 71.27 & 72.18 \\ 
			C & \checkmark & - & 81.12 & 65.67 & 72.24 & 73.01 \\
			D & \checkmark & \checkmark & \textbf{81.42} & \textbf{66.25} & \textbf{73.14} & \textbf{73.60} \\ \hline
		\end{tabular}
	}
	\label{tab_ablation_img}
\end{table}

We perform three sets of ablation experiments on our detection model UniMT.

The Multi-model Mamba Encoder (MME) consists of two complementary branches: ``LiDAR to Image'' (L2I) and ``Image to LiDAR'' (I2L).
We evaluate the contribution of each branch by progressively adding them to a baseline model without MME. 
As shown in Table \ref{tab_ablation_mme}, the baseline achieves an mAP of 71.84\%. 
Incorporating only the L2I branch brings a moderate improvement (+0.86\% mAP), as it enriches image features with geometric context from LiDAR. 
Using only the I2L branch yields a more substantial gain (+1.27\% mAP), highlighting the critical role of projecting rich semantic image features into the 3D space. 
The complete MME, integrating both branches, achieves the highest performance (73.60\% mAP), demonstrating that the dual-space, bidirectional fusion strategy is synergistic and both branches are indispensable for robust multi-modal representation.

We investigate the design of the fusion mechanism in the decoder by comparing our proposed 3D Multi-model Deformable Attention (MDA) with two alternatives: replacing deformable attention with global attention (as in CMT \cite{cmt}), and replacing it with a simple bilinear sampling and concatenation followed by an MLP. 
The results in Table \ref{tab_ablation_mda} clearly show that our MDA module achieves the best performance. 
The global attention variant suffers from a significant performance drop (-1.38\% mAP), as it struggles to focus on precise, informative features from the large-scale BEV and multi-scale image feature maps. 
The bilinear sampling variant performs the worst (-4.69\% mAP), indicating that a naive fusion strategy is insufficient for capturing the complex relationships between object queries and multi-modal features. 
In contrast, MDA efficiently and effectively extracts sparse, relevant features, striking an optimal balance between accuracy and computational cost.

Our model utilizes a pre-trained image backbone and employs a multi-scale feature strategy: the middle-scale feature (16× downsampled) undergoes fusion encoding in the MME, while all three scales (8×, 16×, 32×) are fed into the MDA module. 
We ablate this design along two axes: the use of pre-trained weights and the use of multi-scale features. 
As delineated in Table \ref{tab_ablation_img}, the performance ranks as follows: A \textless ~B \textless ~C \textless ~D. 
Models trained from scratch (A, B) perform worse than their pre-trained counterparts (C, D), underscoring the importance of leveraging robust, general-purpose visual features learned from large-scale datasets. 
Furthermore, for both training strategies, models utilizing multi-scale features (B, D) outperform those using only the fused middle-scale feature (A, C). 
This confirms that while the fused feature is crucial for multi-modal alignment, the original, non-fused features from the pre-trained backbone at different scales preserve unique information that is beneficial for the final decoding stage, and our architecture successfully leverages this hierarchical semantic information.

\subsubsection{\textbf{Ablation Study for Trajectory Prediction Model}}

\begin{table}[t]
	\centering
	\caption{Ablation study on the number of decoder layers in RTMCT}
	\resizebox{85mm}{6mm}{
		\begin{tabular}{cccc}
			\hline
			Decoder Layer Num & ADE(Pedestrian)$_{3/5/10}$ & ADE(Car)$_{3/5/10}$ & ADE(Cyclist)$_{3/5/10}$  \\ \hline
			1 & 0.28/0.25/0.21 & 0.40/0.36/0.32 & 0.87/0.80/0.74 \\
                        2 & \textbf{0.27/0.24/0.21} & \textbf{0.39/0.35/0.31} & \textbf{0.85/0.78/0.72} \\
                        3 & 0.30/0.29/0.26 & 0.40/0.37/0.33 & 0.85/0.79/0.73 \\ \hline
		\end{tabular}
	}
	\label{tab_ablation_traj_layer}
\end{table}

\begin{table}[t]
	\centering
	\caption{Ablation study on class-specific parameterization in RTMCT}
	\resizebox{85mm}{6mm}{
		\begin{tabular}{cccc}
			\hline
			Network Type & ADE(Pedestrian)$_{3/5/10}$ & ADE(Car)$_{3/5/10}$ & ADE(Cyclist)$_{3/5/10}$  \\ \hline
			All Shared & 0.28/0.25/0.22 & 0.45/0.41/0.38 & 0.87/0.80/0.74 \\
                        All Specific & 0.27/0.24/0.21 & 0.40/0.37/0.33 & 0.90/0.82/0.77 \\
                        Original Network & \textbf{0.27/0.24/0.21} & \textbf{0.39/0.35/0.31} & \textbf{0.85/0.78/0.72} \\ \hline
		\end{tabular}
	}
	\label{tab_ablation_traj_class}
\end{table}

We conduct two ablation studies on our trajectory prediction model RTMCT.

We experiment with varying the number of decoder layers in the Transformer. 
As shown in Table \ref{tab_ablation_traj_layer}, using a single decoder layer results in suboptimal performance, as the model lacks sufficient capacity for complex interaction modeling between the ego-trajectory and its neighbors. 
Increasing the depth to two layers yields a performance boost, indicating an improved ability to reason about social contexts. 
However, further increasing to three layers leads to a slight degradation in performance, suggesting potential overfitting on the training data. 
Therefore, we adopt two decoder layers as the optimal configuration for our model.

RTMCT employs a hybrid parameterization strategy: using class-specific linear layers for trajectory encoding and the prediction head, while using shared parameters in the Transformer decoder. 
We compare this strategy with two alternatives: using class-specific parameters throughout the entire network ("All Specific"), and using shared parameters for all classes across the entire network ("All Shared"). 
The results in Table \ref{tab_ablation_traj_class} demonstrate that our hybrid strategy achieves the best balance. 
The "All Shared" model performs worst, as it fails to capture the distinct motion patterns and influences of different object classes. 
The "All Specific" model, while capturing class-specific traits, likely suffers from over-parameterization and reduced generalization due to the limited data for certain classes (e.g., Cyclist). 
Our hybrid approach effectively leverages class-specific encoding and decoding to model intrinsic class behaviors, while using a shared interaction module in the decoder to learn a common, efficient social reasoning mechanism, resulting in the most robust and accurate predictions.

\section{CONCLUSION}

This paper has presented an efficient multi-modal perception framework for resource-constrained service mobile robots, enabling real-time 3D detection and trajectory prediction of dynamic objects such as pedestrians, vehicles, and cyclists. 
To overcome the limitations of existing methods, we introduced two novel models: the Unified modality detector with Mamba and Transformer (UniMT) for detection and the Reference Trajectory-based Multi-Class Transformer (RTMCT) for trajectory prediction.

Our UniMT detector achieves an superior balance between accuracy and efficiency through its novel fusion strategy. 
The Multi-model Mamba Encoder (MME) performs deep, bidirectional feature fusion in both image and 3D space with linear complexity, while the 3D Multi-model Deformable Attention (MDA) enables the decoder to extract precise features from multi-modal inputs efficiently. 
Extensive evaluations on both the CODa benchmark and the challenging nuScenes detection benchmark demonstrate that UniMT outperforms several strong baselines, achieving competitive detection accuracy with reduced inference time. 
For trajectory prediction, RTMCT effectively handles flexible-length, multi-class historical trajectories and generates diverse future predictions without complex generative models. 
By leveraging learnable reference trajectories and a simple Transformer architecture, it achieves significant improvements in prediction accuracy with remarkably fast inference.

The practical deployability of our framework has been successfully demonstrated through implementation on a wheelchair robot equipped with an entry-level GPU. 
With minimal fine-tuning, the system achieved real-time performance at 13.9 FPS while maintaining satisfactory accuracy. 
This successful transfer, coupled with our released code and ROS implementation, underscores the practical value and generalizability of our work.

However, the performance of our system requires further improvement to handle more complex scenarios. 
Several strategies can be adopted to enhance the model architecture, aiming to boost detection and trajectory prediction accuracy while preserving its lightweight nature. 
Currently, our detection model, UniMT, relies solely on single-frame sensor data, limiting its capacity to leverage temporal context for addressing occlusions or noisy detections. 
Future versions could integrate features or object queries from previous frames to exploit temporal consistency. 
Similarly, our trajectory prediction model, RTMCT, does not yet incorporate environmental cues—such as geometric layouts or semantic maps—which are essential for accurately forecasting motion in structured scenes. 
Enriching the model with such contextual information is expected to substantially reduce prediction errors.

\section*{ACKNOWLEDGMENT}
This work is supported by the National Natural Science Foundation of China (Grant 62573287), and the Science and Technology Commission of Shanghai Municipality (Grant 20DZ2220400).

\bibliographystyle{IEEEtran}
\bibliography{IEEEabrv, references}

% \newpage

 

% \end{IEEEbiography}



% \vfill

\end{document}